\documentclass{svproc}
\usepackage[utf8]{inputenc}

\usepackage[T1]{fontenc}
\usepackage{newtxtext}
\usepackage{newtxmath}

\usepackage{url}
\usepackage{booktabs}
\usepackage{array}
\usepackage{amsmath}
\usepackage{graphicx}
\usepackage{siunitx}
\usepackage{microtype}
\usepackage{placeins}
\usepackage{float}

\graphicspath{{figures/}}

\AtBeginDocument{%
  \ifx\pdfinfo\undefined
  \else
    \pdfinfo{%
      /Title (Auditable Retrieval-Triggered Projection Correction for Frozen VLA Control: An Empirical and Failure Analysis)

      /Subject (Empirical analysis of an auditable external action correction for frozen vision-language-action control)
      /Keywords (vision-language-action model; external intervention; action correction; failure analysis; CALVIN; robot manipulation)
    }%
  \fi
}

\begin{document}
\mainmatter

\title{Retrieval-triggered Action Projection Adjustment\\for Frozen VLA Policies: An Empirical Study and Failure Analysis}
\titlerunning{ }

\author{Jinming Wang \quad \inst{[0009-0000-0988-1172]}, Ruiheng Zhang \quad \inst{[0009-0004-5393-4656]}, Yong Wang \quad  \inst{[0009-0003-2943-4670]}, Zipan Xiong \quad\inst{[0009-0006-1279-0179]}, Keyuan Luo \quad \inst{*[0009-0006-5265-784\mathrm{X}]}}
\authorrunning{ }

  \institute{Department of Computer Science and Technology, Hainan Bielefeld University of Applied Sciences, Danzhou, China\\
\email{ keyuan.luo@hibiuh.edu.cn}}

\maketitle

\begingroup
\renewcommand{\thefootnote}{}
\footnotetext{\hspace*{-1em}Jinming Wang and Ruiheng Zhang contributed equally to this work.}
\endgroup

\begin{abstract}
Frozen Vision-Language-Action (VLA) policies cannot incorporate previously encountered failures without updating the underlying model. We present retrieval-triggered projection correction (RTPC), an auditable external-memory wrapper for frozen OpenVLA policies that retrieves stress-collected local-depth records and applies translation-only projection correction while keeping rotation and gripper commands unchanged. RTPC is evaluated on CALVIN single-task manipulation using primary and independent replication cohorts. In the primary evaluation, RTPC reduced a depth-based near-collision or stall proxy event rate without changing task-success count. However, this reduction was not reproduced in the independent replication, and retrieved correction directions were not distinguishable from random same-family directions. Analysis identifies endpoint coupling, batch sensitivity, and the lack of a trigger-matched memory-free baseline as limitations for attributing effects to retrieval-based correction. These findings do not establish physical safety improvement or memory-specific benefits, but provide an empirical evaluation and failure analysis of retrieval-conditioned intervention for frozen VLA policies.
\end{abstract}

\section{Introduction}

Vision-Language-Action (VLA) models map visual observations and language instructions to robot actions by transferring large-scale visual and linguistic knowledge to robotic control~\cite{driess_palm-e_2023,li_vision-language_2024,al_rt-2_2023}. OpenVLA offers an open pretrained policy, while CALVIN provides language-conditioned manipulation tasks with a standardized policy interface~\cite{openvla,calvin}. A frozen policy does not retain information about prior adverse observations. Prior autonomous systems have explored persistent external information storage to support reliability-oriented decision processes~\cite{li2023shuttle}. An external memory therefore provides an inspectable alternative to policy retraining or architectural modification. We ask whether a simple external correction can produce a reproducible
and interpretable effect. RTPC mines stress episodes using gripper-depth,
translational-motion, and low tool-center-point (TCP) progress cues to
store local-depth descriptors and correction directions. During
evaluation, same-family retrieval gates a one-step projection correction
around frozen OpenVLA. The endpoint is a depth-proximity or stall proxy,
not physical contact or a certified safety condition. Across eight
single tasks, a 112-state primary batch and an independent 40-state
replication batch test proxy change, task success, replication, and
retrieved-versus-random directions. 
This paper makes the following contributions:
{
\renewcommand{\labelitemi}{\textbullet}
\begin{itemize}
    \setlength{\topsep}{2pt}
    \setlength{\itemsep}{1pt}
    \setlength{\parsep}{0pt}
    \setlength{\parskip}{0pt}

    \item We present RTPC, an auditable external-memory wrapper that
    retrieves stress-collected local-depth records and applies
    translation-only projection correction to actions proposed by a
    frozen OpenVLA policy.

    \item We show that the primary proxy effect did not replicate,
    task success did not improve, and retrieved directions were not
    statistically distinguishable from random same-family directions.

    \item We identify endpoint coupling as a barrier to physical-safety
    interpretation, and batch sensitivity and the absence of a
    trigger-matched, memory-free baseline as barriers to memory-specific
    attribution.
\end{itemize}
}

\begin{figure}[!t]
\centering
\includegraphics[width=\textwidth]{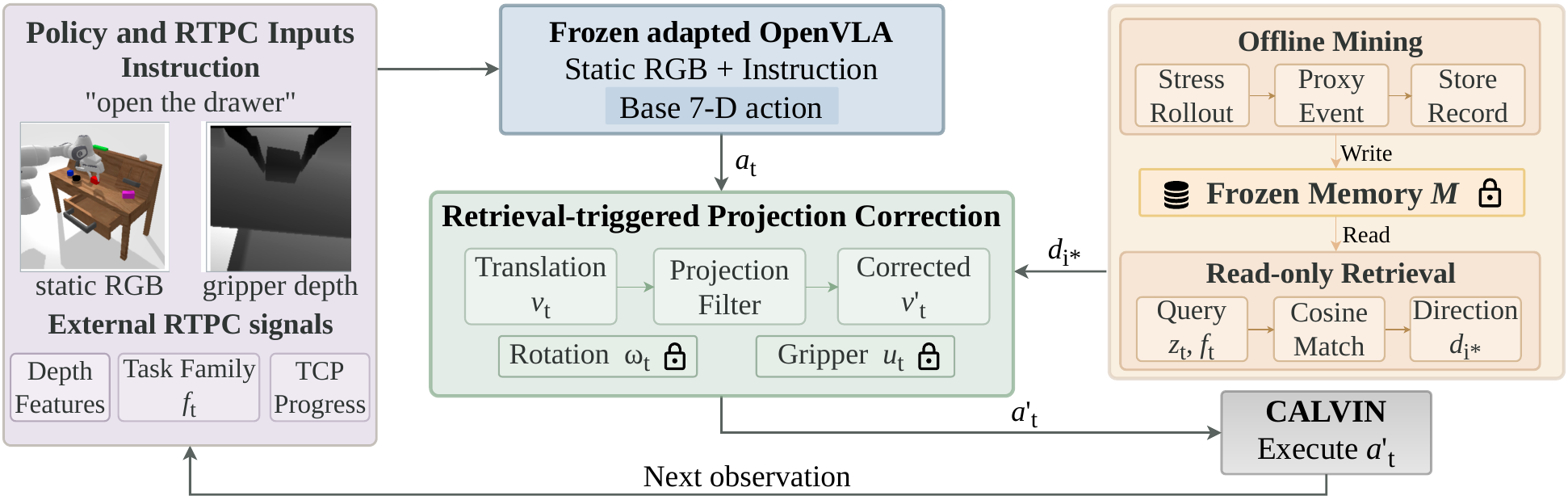}
\caption{\makebox[\textwidth][l]{RTPC workflow in CALVIN}}
\label{fig:method}
\end{figure}

\section{Related Work}
Multimodal transformers and cross-modal attention have been applied to pedestrian crossing-intention prediction in safety-critical urban settings, while diffusion-based visuomotor controllers have demonstrated strong manipulation performance~\cite{chi_diffusion_2023,li_acit_2025,li_mtc_2025,li_multicontext_2025}. OpenVLA is an open 7B-parameter vision-language-action model that supports parameter-efficient adaptation~\cite{ghosh_octo_2024,openvla,open_x-embodiment_collaboration_open_2024}. Recent VLA memory systems have investigated recurrent history representations, experience retrieval, and agent-level reuse of previous interactions~\cite{muvla,physmem,lamemvla,halo,dejavu,harnessvla}. These approaches incorporate prior experience through learned representations, retrieved context, or agent-level memory mechanisms, following the broader idea of retrieval-augmented reasoning~\cite{borgeaud_improving_2022,lewis_retrieval-augmented_2020}. In contrast, RTPC maintains a frozen external correction memory and applies a deterministic one-step modification to the translational action without updating the underlying policy.

Shielding, control barrier functions, and recovery policies constrain or replace potentially unsafe actions using specified dynamics, formal constraints, or learned risk estimates~\cite{alshiekh,ames,dawson,recoveryrl}. SafeVLA instead modifies a VLA policy through constrained learning and explicit safety costs~\cite{safevla}. RTPC differs from these methods because it applies retrieval-conditioned projection correction around a frozen policy. It does not optimize a safety objective, use a contact-grounded endpoint, or provide an invariance guarantee. The measured proxy events should therefore not be interpreted as evidence of formal or physical safety.

\section{Method}

\subsection{Policy and Proxy Event}

At time $t$, the adapted policy predicts a seven-dimensional relative action
\begin{equation}
 \vec{a}_t=\pi_{\mathrm{VLA}}(I_t,g)
 =[\vec{v}_t,\vec{\omega}_t,u_t]\,\,,
\end{equation}
where $I_t$ is the static red-green-blue (RGB) image, $g$ is the instruction, $\vec{v}_t\in\mathbb{R}^{3}$ is translation, $\vec{\omega}_t\in\mathbb{R}^{3}$ is rotation, and $u_t$ is the gripper command. All conditions use the same 1st--99th percentile clipping bounds derived from the training actions and the same CALVIN gripper binarization.

The side channel uses an $84\times84$ metric depth image from the gripper camera.
Within the crop $D_t[10{:}74,\,10{:}74]$, $q_t^{05}$ denotes the
fifth percentile of the valid depth values. The fixed threshold is
$\tau_d=0.104\,\mathrm{m}$. It is defined as the larger of
$0.04\,\mathrm{m}$ and the second percentile of $q^{05}$ computed
over 2,000 fixed training frames.

A five-step window is a candidate when at least three frames satisfy $q_t^{05}<\tau_d$, depth decreases by at least $0.005\,\mathrm{m}$ relative to the preceding three-frame median, and $\lVert\vec{v}_t\rVert_2\geq0.005$. It must also contain at least $0.02\,\mathrm{m}$ cumulative commanded translation while realized TCP displacement is below $0.01\,\mathrm{m}$. If TCP state is unavailable, the last three $q^{05}$ values must span at most $0.005\,\mathrm{m}$. Event detection is suppressed for ten steps, and the detector state resets each episode. Because commanded translation is both corrected and tested, RTPC can render a candidate window ineligible. Consequently, the endpoint is not intervention-independent.

\subsection{Stress-based Proxy-event Memory Construction}

\begin{figure}[t]
\centering
\includegraphics[width=\textwidth]{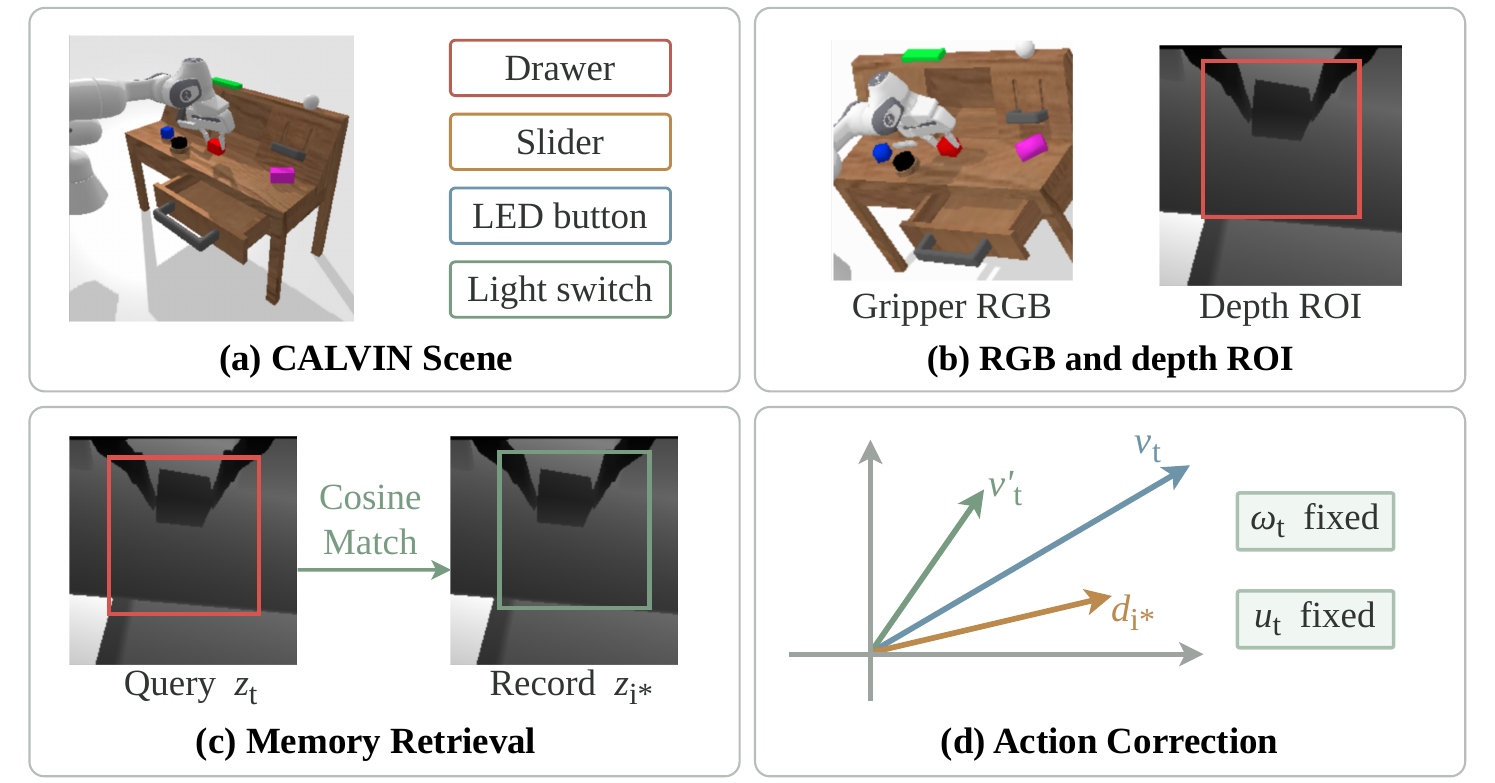}
\caption{\makebox[\textwidth][l]{CALVIN observations, memory retrieval, and action correction}}
\label{fig:setup}
\end{figure}

The memory was constructed from 64 rollouts, evenly divided between natural and stress conditions. The natural-condition rollouts generated no writable events. One stress-mining pass produced 721 stress steps, with at most three consecutive stress actions. Seven entries were retained, including three drawer records, two light-switch records, one slider record, and one light-emitting diode (LED) button record. The target of eight entries, with two per family, was not met, and all retained entries were collected under stress conditions. The fixture families and synchronized gripper RGB and depth observations are shown in panels (a) and (b) of Fig.~\ref{fig:setup}. Same-family retrieval from the frozen correction memory and the resulting translational correction are illustrated in panels (c) and (d), respectively, with rotation and gripper commands unchanged.

Invalid depth measurements and values below $0.02\,\mathrm{m}$ are replaced with $1.0\,\mathrm{m}$. The resulting depth values are then clipped to $[0.03,1.0]\,\mathrm{m}$. Average pooling reduces the $64\times64$ crop to 256 proximity features. These features are concatenated with a 16-bin normalized histogram and six summary statistics, producing a 278-dimensional descriptor. The summary statistics comprise $q^{05}$, $q^{25}$, $q^{50}$, $q^{75}$, the mean, and the standard deviation. Standardization using 2,000 training frames, followed by $\ell_2$ normalization, yields the descriptor $\vec{z}_i$. Each memory entry stores $\vec{z}_i$, the task family $f_i$, its provenance, and a numerically stabilized translation direction defined as
\begin{equation}
\vec{d}_i =
\frac{\vec{v}_i}{\lVert\vec{v}_i\rVert_2+\epsilon}\,\,,
\label{eq:stored-direction}
\end{equation}
where $\vec{v}_i$ denotes the translational command associated with the retained entry and $\epsilon$ is a small positive constant used for numerical stability. The vector $\vec{d}_i$ is stored as the correction direction for retrieval-conditioned action modification.

\subsection{Retrieval and Projection Correction}

During evaluation, both the policy and correction memory remain frozen, and RTPC uses same-family depth retrieval to determine whether the proposed translation should be corrected. Retrieval is enabled when $q_t^{05}\leq\tau_d+0.03\,\mathrm{m}$. Instructions are mapped deterministically to four fixture families. Same-family entries are ranked by $s(\vec{z}_t,\vec{z}_i)=\vec{z}_t^\top\vec{z}_i$. The best entry is returned when $s\geq\tau$. Otherwise, the base action is retained. The build-data threshold $\tau=0.470$ used a nominal 20\% false-trigger limit. The strict and permissive sensitivity values are 0.651 and 0.319, respectively. The false-trigger labeling rule was not retained, preventing independent reconstruction of this calibration. Given a retrieved direction $\vec{d}_i$, RTPC applies
\begin{equation}
\vec{v}'_t=\vec{v}_t-
\max(0,\vec{v}_t^\top\vec{d}_i)\vec{d}_i\,\,.
\label{eq:projection}
\end{equation}
If $q_t^{05}<0.8\tau_d$, $\vec{v}'_t$ is multiplied by 0.5. Rotation and gripper commands remain unchanged, and histories reset each episode. Correction precedes common clipping and binarization. Under the correction defined in Equation~\ref{eq:projection}, translational motion orthogonal to $\vec{d}_i$ is preserved, but its safety is not evaluated.

\section{Experimental Protocol}

\subsection{Policy Adaptation}

The evaluated policy is OpenVLA 7B with the quick\_500 adapter, which was trained for 500 QLoRA steps on the CALVIN training split to map static RGB observations and language annotations to seven-dimensional relative Cartesian actions. Evaluation uses deterministic decoding, percentile clipping, and gripper binarization. Because the archived record does not preserve the complete training configuration, the conclusions are restricted to this checkpoint.

\subsection{Evaluation Design}

Eight single tasks from the task\_ABC\_D tree~\cite{calvinrepo} cover
drawer, slider, LED-button, and light-switch operations, with each
episode limited to 180 steps. The amended protocol removed annotation
indices, start frames, and initial-state hashes used in earlier builds
or tests. Candidate scarcity reduced the allocation from 20 to 14
initial states per task, leaving 112 paired primary states with no
identifier-level overlap in the exclusion keys. Overlapping annotation
intervals grouped these states into 14 connected components, so they
are not fully independent at the trajectory level. A separate 40-state
replication batch with 10 states per fixture family passed the same
checks, and the pooled 152-state analysis remained post hoc. Retained metadata identify a held-out CALVIN validation environment but not its A/B/C/D label. Therefore, no claims are made about five-subtask challenge performance or cross-scene transfer.

The four paired conditions share the same checkpoint and initial states.
Here, $B_0$ is unmodified, $B_1$ halves translation in the proximity
region, $B_2$ applies the $B_3$ gate and projection with a random
same-family direction, and $B_3$ retrieves the highest-similarity
direction. In the primary batch, $B_1$, $B_2$, and $B_3$ modified
34.46\%, 7.56\%, and 6.38\% of steps, respectively, showing that
$B_1$ was not trigger-matched to $B_3$. The 95\% confidence interval
for the proxy-rate difference between $B_3$ and $B_2$ included zero,
while the family gate provided task-label information. Without a
trigger-matched memory-free depth baseline, persistent-memory effects
cannot be isolated. Evaluation included 3,168 rollouts and 27.92
effective GPU-hours across primary, replication and sensitivity
conditions.

\subsection{Metrics and Statistics}

For episode $j$ with $E_j$ events and $T_j$ executed steps, the primary proxy rate gives each initial-state pair equal weight and is defined as
\begin{equation}
 \bar r=\frac{1}{N}\sum_{j=1}^{N}\frac{100E_j}{T_j}\,\,.
\end{equation}
This avoids weighting long failed episodes more heavily after successful episodes terminate. The pooled rate, $100\tfrac{\sum_j E_j}{\sum_j T_j}$, describes exposure per step. We also report success, event-bearing episodes, repeated-event episodes, trigger rate, and latency. Primary intervals use 10,000 task-family-stratified paired-bootstrap samples. An interval-component cluster bootstrap tests dependence, and exact McNemar tests summarize binary flips.

Event counts, incidence, and repetition are derived from the same action-coupled detector. Contact, force, minimum realized $q^{05}$, time below $\tau_d$, realized TCP series, and threshold-crossing counts are unavailable. Threshold, shuffle, family, and combined analyses are exploratory and unadjusted for multiplicity.

After examining the earlier results, we set a post hoc success tolerance 10
percentage points below the 22.3\% success rate under $B_0$. This threshold is
a permissive engineering screen rather than a confirmatory non-inferiority margin
and does not establish preservation of policy competence.

\begin{table}[!t]
\caption{Primary batch results}
\label{tab:primary}
\centering
\small
\renewcommand{\arraystretch}{1.12}
\begin{tabular*}{\textwidth}{
  @{\extracolsep{\fill}}
  c c c c c c c
  @{}
}
\toprule
Condition &
Success &
Events &
Steps &
Mean rate &
Pooled rate &
Trigger (\%) \\
\midrule

$B_0$ Base &
25 (22.3\%) &
10 &
17,658 &
0.0496 &
0.0566 &
0.00 \\

$B_1$ Slow &
19 (17.0\%) &
11 &
18,163 &
0.0546 &
0.0606 &
34.46 \\

$B_2$ Random &
22 (19.6\%) &
6 &
17,722 &
0.0298 &
0.0339 &
7.56 \\

\textbf{$B_3$ RTPC} &
\textbf{25 (22.3\%)} &
\textbf{2} &
\textbf{17,270} &
\textbf{0.0099} &
\textbf{0.0116} &
\textbf{6.38} \\
\bottomrule
\end{tabular*}
\end{table}

\section{Results}

\subsection{Amended Primary Batch}

In the primary batch, $B_3$ produced two proxy events, compared with ten under $B_0$,
while both conditions achieved 25 successes in 112 trials, as summarized in
Table~\ref{tab:primary}. The episode-mean difference between
$B_3$ and $B_0$ was $-0.0397$, with a 95\% confidence interval from $-0.0694$ to
$-0.0099$. The interval-clustered range was $-0.0676$ to $-0.0101$. The pooled proxy
rates were 0.0116 and 0.0566, respectively. Events occurred in 2 of 112 $B_3$ episodes
and 10 of 112 $B_0$ episodes, with one $B_3$-only and nine $B_0$-only cases.
McNemar's exact two-sided test yielded $p=0.0215$. Because this analysis uses the same
detector endpoint, it does not independently establish reduced physical risk.

Task success was identical under $B_3$ and $B_0$, giving a difference
of $0.0$ percentage points and a 95\% confidence interval from $-6.25$
to $6.25$. Seven paired outcomes changed in each direction, and
McNemar's test gave $p=1.0$. The proxy difference between $B_3$ and
$B_2$ was $-0.0198$, with a 95\% confidence interval from $-0.0496$
to $0.0050$. Because the interval included zero, retrieved directions
were not statistically distinguishable from random same-family directions. No repeated-event episodes occurred under any condition. Moreover, only ten pairs were discordant in the event-incidence comparison, so the estimate remains sensitive to changes in a small number of episodes. The mean rate is episode-weighted, whereas the pooled rate is computed from raw event and step counts. The $B_1$ and $B_3$ conditions are not budget-matched.

\subsection{Independent Replication and Combined Analysis}

The 40-state replication did not reproduce the proxy change. $B_0$,
$B_2$, and $B_3$ each had a rate of 0.0694. The difference between
$B_3$ and $B_0$ was 0.0000, with a 95\% confidence interval from
$-0.0556$ to $0.0556$. The task-success difference was $-2.5$
percentage points, with a 95\% confidence interval from $-10.0$ to
$5.0$. 

\begin{table}[t]
\caption{Batch-wise effects of $B_3$ relative to $B_0$}
\label{tab:batches}
\centering
\small
\renewcommand{\arraystretch}{1.12}
\begin{tabular*}{\textwidth}{
  @{\extracolsep{\fill}}
  c c c c c c
  @{}
}
\toprule
Outcome &
Batch &
$N$ &
Difference &
Lower &
Upper \\
\midrule
&
Primary &
112 &
$\textbf{--0.0397}$ &
$\textbf{--0.0694}$ &
$\textbf{--0.0099}$ \\

Proxy &
Replication &
40 &
$\textbf{0.0000}$ &
$-0.0556$ &
$0.0556$ \\

&
Combined &
152 &
$-0.0292$ &
$-0.0585$ &
$-0.0037$ \\

\addlinespace[2pt]

&
Primary &
112 &
$0.00$ &
$-6.25$ &
$6.25$ \\

Success &
Replication &
40 &
$-2.50$ &
$-10.00$ &
$5.00$ \\

&
Combined &
152 &
$-0.66$ &
$-5.92$ &
$4.61$ \\
\bottomrule
\end{tabular*}
\end{table}

\FloatBarrier

\begin{figure}[!htbp]
\centering
\includegraphics[width=\textwidth]{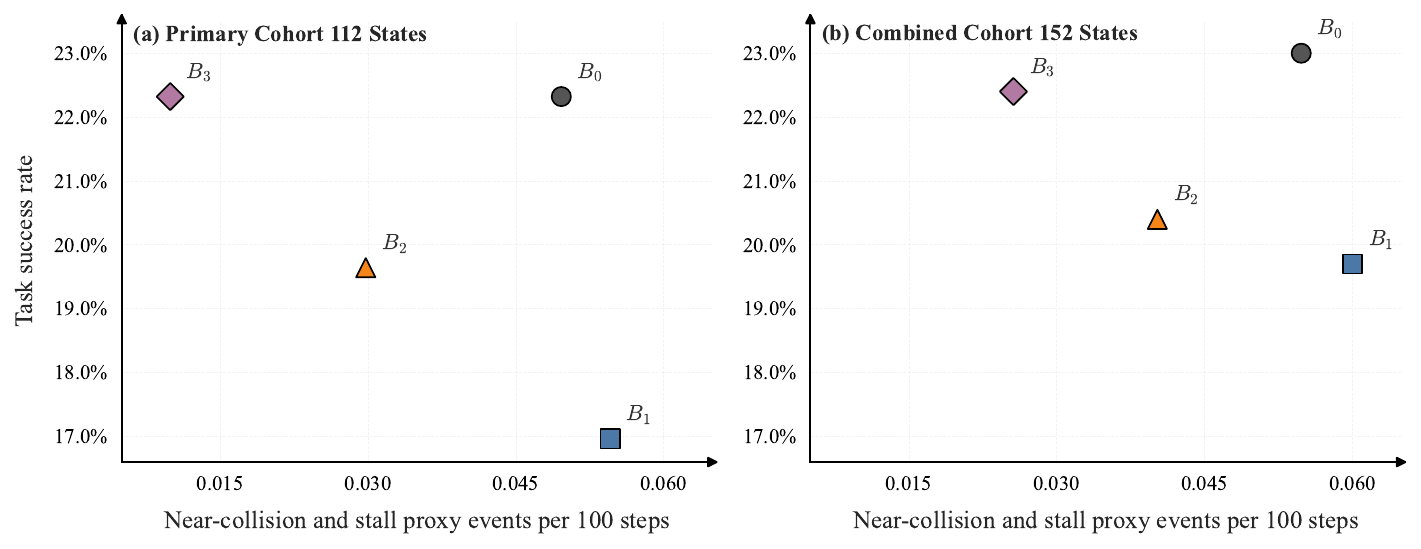}
\caption{\makebox[\textwidth][l]{Proxy-rate and task-success estimates by cohort}}
\label{fig:cohort_tradeoff}
\end{figure}

The post hoc analysis of the pooled 152 states favored $B_3$ on the
proxy but not on task success, while repeated-event rates were identical
across methods. Because the effects differed between the 112-state
primary batch and the 40-state replication batch, the pooled result
should be treated as exploratory rather than as an independent
replication or confirmatory finding. Cohort-level proxy and task-success
estimates are shown in Fig.~\ref{fig:cohort_tradeoff}, and the
corresponding 95\% confidence intervals for the differences between
$B_3$ and $B_0$ are reported in Table~\ref{tab:batches}.

\FloatBarrier

\subsection{Sensitivity and Mechanism Analysis}

RTPC was tested across retrieval thresholds, direction variants, and
task-family settings. Proxy rates changed non-monotonically with the
threshold. Under the strict, base, and permissive settings, $B_3$
reached rates of 0.0446, 0.0099, and 0.0248, with 23, 25, and 21
successes, respectively. The corresponding $B_2$ rates were 0.0496,
0.0298, and 0.0198. The natural-only condition matched $B_0$ because
none of its 32 construction rollouts produced a memory entry. These
exploratory results revealed no clear persistent-memory benefit.

A robustness analysis defined before outcome inspection produced rates
of 0.0446, 0.0397, and 0.0550 for the action-shuffled,
descriptor-shuffled, and family-only settings. The associated contrasts
with $B_3$ often included zero. A separate post hoc analysis applied
three global shuffles. Action rates ranged from 0.0198 to 0.0496, and
descriptor rates ranged from 0.0287 to 0.0446. This variation leaves
the underlying mechanism unresolved.

Effects also varied across task families. A positive descriptive proxy
difference appeared only for the drawer family. The light-switch family
had no proxy events, yet task success fell from 10.7\% to 3.6\%.
Leave-one-family-out differences remained nonpositive, with two
confidence intervals having upper bounds of zero. No single family
drove the primary pattern, and the effect was not consistent across
families.

\subsection{Latency}

Total latency was $725.22 \pm 5.12\,\mathrm{ms}$ under $B_0$ and $727.37 \pm 6.83\,\mathrm{ms}$ under $B_3$, reported as mean and standard deviation. The $2.15\,\mathrm{ms}$ increase mainly reflected variation in policy time, while retrieval averaged $1.183\,\mathrm{ms}$. These measurements are platform-specific because latency depends on the execution setup.

\section{Discussion and Limitations}\label{sec:limitations}

The primary reduction in the proxy did not replicate, task success
remained unchanged, and $B_3$ was statistically indistinguishable from
$B_2$. These findings support viewing the study as a stress test and
failure analysis rather than as validation of a memory-based method.
Interpretation is further constrained because the same depth-derived
statistic drives record construction, retrieval gating, and proxy
evaluation, while commanded translation influences both event
eligibility and correction. Fewer events may thus arise from altered
trajectories, detector interactions, or both. Without independent
physical outcomes or a trigger-matched memory-free baseline, the
results do not demonstrate reduced physical risk or a memory-specific
benefit.

The evaluation is limited to one frozen checkpoint, stress-only memory,
incomplete family coverage, and deterministic single-task CALVIN.
Incomplete records of mining and adaptation prevent exact reproduction,
while performance across scenes, policies, and real robots remains
unknown. Retrieved directions may also become outdated after system
changes, and RTPC provides no formal safety guarantee.

\section{Conclusion}

This study evaluates an auditable external-memory correction framework
around a single frozen OpenVLA checkpoint. The primary proxy reduction
did not replicate, task success remained unchanged, and retrieved
directions were statistically indistinguishable from random same-family
directions. Since the endpoint depends on the commanded action, the
findings cannot be interpreted as evidence of reduced physical risk.
Further evaluation should use intervention-independent endpoints,
trigger-matched memory-free baselines, and additional seeds and checkpoints
to assess batch sensitivity.

\bibliographystyle{spmpsci}
\bibliography{references}

\end{document}